\documentclass[10pt,twocolumn]{ICCAS}

\usepackage{graphicx}
\usepackage{graphics}
\usepackage{epsfig}

\usepackage{amsbsy}
\usepackage{bm}

\usepackage{cite}
\usepackage{booktabs}
\usepackage{multirow}
\usepackage{tabularx}
\usepackage{siunitx}

\usepackage{xcolor}
\usepackage{textcomp}
\usepackage{gensymb}
\usepackage{caption}
\usepackage{subcaption}

\usepackage{tikz}
\usetikzlibrary{arrows.meta, positioning, calc}

\usepackage{algorithm}
\usepackage{algorithmic}

\usepackage{microtype}
\usepackage{verbatim}
\usepackage{soul}
\usepackage{ragged2e}
\usepackage{placeins}
\usepackage{dblfloatfix}
\usepackage{float}

\usepackage[autolanguage]{numprint}

\usepackage[colorlinks=true,linkcolor=blue,citecolor=blue,urlcolor=blue]{hyperref}

\def\BibTeX{{\rm B\kern-.05em{\sc i\kern-.025em b}\kern-.08em
    T\kern-.1667em\lower.7ex\hbox{E}\kern-.125emX}}

\begin{document}

\title{Hybrid Attention Estimation Pipeline for Adaptive HRI Using an Expressive Robotic Head}

\author{
Pablo Moraes${}^{1}$,
Mónica Rodríguez${}^{1}$,
Christopher Peters${}^{2}$,
Hiago Sodre${}^{1}$,
Tobias Doernbach${}^{2}$,
Bruna Guterres${}^{1}$,
and Ricardo Grando${}^{1*}$
}

\affils{
${}^{1}$Technological University of Uruguay, Uruguay\\
${}^{2}$Ostfalia University of Applied Sciences, Germany\\
{\small ${}^{*}$Corresponding author: pablo.moraes@utec.edu.uy}
}

\abstract{
This paper presents an applied case study on hybrid visual attention estimation for human--robot interaction using an expressive robotic head based on the InMoov ecosystem. The proposed pipeline combines a fast geometric perception layer with an independent semantic perception layer based on a vision--language model. The geometric layer provides high-frequency face and head-pose information for temporal regulation, while the semantic layer receives only raw egocentric camera frames and produces contextual attention labels related to attention toward the robot, phone use, or attention elsewhere. These signals are integrated through a finite state machine that regulates adaptive interaction behavior, including activation, waiting, interaction resumption, and return to rest. The system was evaluated with 10 participants across 40 trials covering baseline and adaptive interaction conditions. Results show reliable interaction start across all trials, consistent pause behavior in the adaptive distraction condition, and non-redundant semantic information between the geometric and semantic outputs. A demonstration video of the system is available at: \url{https://youtube.com/shorts/jbKdYOb6ezs?si=JfecAN6WGFC9BrQH}.
}


\maketitle

\section{Introduction}

In social contexts, human--robot interaction (HRI) requires robots to generate verbal responses and to perceive non-verbal user cues, adapting their behavior in a visible and coherent manner. Visual attention is a key cue that enables the system to infer user availability, intention, and engagement. For robots with expressive heads, head orientation, eye contact, and facial expressions are particularly important channels for conveying the system's internal state and willingness to interact \cite{Admoni2017SocialEyeGaze,RojasQuintero2021SensorHeads}.

It has been reported that mechanisms such as eye contact and joint attention play a structural role in social interaction, fostering coordination between human and robot and improving the fluency of the exchange \cite{Kozima2003AttentionCoupling,Staudte2009VisualAttention,Huang2011JointAttention}. Likewise, visual attention estimation has been addressed through cues such as face detection, head pose, body orientation, gaze direction, and contextual information, showing that attention should be understood as a dynamic and multimodal state rather than as an isolated binary variable \cite{Li2012VisionBasedAttention,Veronese2017ProbabilisticMapping,Das2012VisionBasedAttentionControl,Begum2009ProbabilisticAttention}.

However, much of the literature addresses attention estimation, expressive cue design, and physical platform implementation separately. Although accessible HRI platforms and low-cost expressive robotic heads have been proposed, a gap still remains between visual attention perception and its systematic translation into embodied expressive behaviors within a reproducible and physically accessible robotic head \cite{Berra2019Berrick,Faraj2020FaciallyExpressive,Netzev2019ManyFacedRobot}. In this context, the InMoov ecosystem is particularly relevant as an open and reproducible basis for the development of humanoid and animatronic platforms oriented toward experimentation\cite{Langevin2012InMoov,peters2025remote}. In addition, although vision--language models offer new possibilities for interpreting visual context in social robots, their integration into embodied real-time HRI systems remains challenging \cite{Janssens2025MultimodalSocialConversations}.

\begin{figure}[t]
    \centering
    \includegraphics[width=0.75\linewidth]{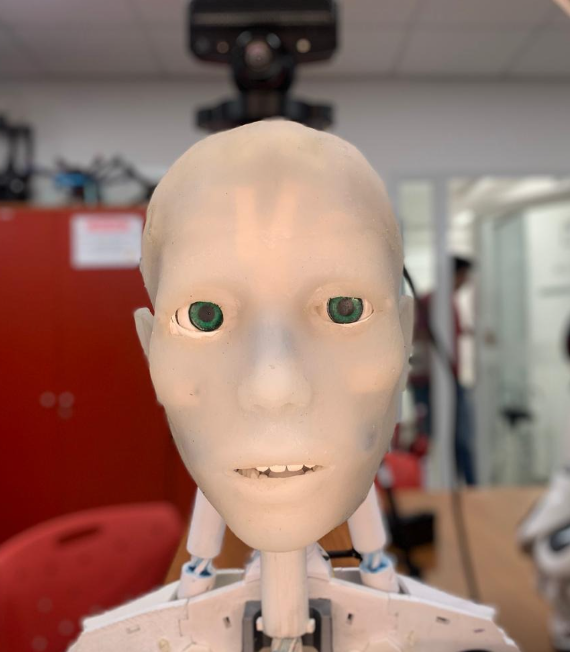}
    \caption{Expressive robotic head based on the InMoov open-source platform.}
    \label{fig:inmoov_platform}
\end{figure}

This work presents an applied case study on hybrid visual attention estimation for HRI using an expressive robotic head based on InMoov. The proposed pipeline combines a fast geometric perception layer for user presence and attention estimation with an independent semantic layer for contextual interpretation. Based on this information, the system generates discrete social states that regulate observable robot behaviors such as activation, waiting, resuming, and returning to rest.

The main contribution of this work is a hybrid attention-estimation pipeline that connects visual attention estimation, contextual interpretation, and adaptive interaction behavior in a physical robotic platform. The proposed system improves the interaction pipeline by combining a high-frequency geometric attention layer for real-time state regulation with an independent VLM-based semantic observer, enabling attention-aware pause and resume behaviors during human--robot interaction. The pipeline is evaluated through objective metrics of interaction start reliability, protocol outcomes, response timing, and geometric--semantic divergence. To support reproducibility, \textbf{the open-source implementation is freely available on \url{https://github.com/ShiiSuii/ICCAS_STUDY}}.

\section{Related Work}

Visual attention and social gaze have been widely studied as central components of human--robot interaction. Joint attention mechanisms allow humans and robots to coordinate their focus during interaction, improving task performance, social presence, and perceived robot competence \cite{Huang2011JointAttention,Domhof2015MultimodalJointAttention,Pereira2019ResponsiveJointAttention}. More broadly, Admoni et al. \cite{Admoni2017SocialEyeGaze} organize social gaze in HRI from human-centered, design-centered, and technology-centered perspectives, highlighting its relevance both as a perceptual cue and as an expressive behavior.

Several works have addressed visual attention estimation from observable user cues such as face detection, head pose, gaze direction, body orientation, and facial expression. Veronese et al. \cite{Veronese2017ProbabilisticMapping} showed that head pose can be used as a practical signal for estimating attentional direction, while Li et al. \cite{Li2012VisionBasedAttention} proposed a vision-based system that combines multiple cues for natural interaction in open environments. Attention has also been used as a control variable for robot behavior. Das et al. \cite{Das2012VisionBasedAttentionControl,Das2013AttentionControl} proposed vision-based attention control mechanisms that regulate awareness signals, eye contact, and attentional redirection. Similarly, Oishi et al. \cite{Oishi2016RoboticAttendant} modeled robot behavior through a finite state machine adapted to the perceived state of the person. These works are closely related to the present study, since attention is not only estimated, but also used to regulate visible interaction states.

Expressive robotic heads provide suitable physical platforms for studying this type of interaction, since they combine perception, head orientation, and facial expressiveness in a compact embodied system. Rojas-Quintero et al. \cite{RojasQuintero2021SensorHeads} review sensor heads for humanoid robots and distinguish between expressive and non-expressive platforms. Other works have proposed accessible or open robotic heads for HRI, including Berrick \cite{Berra2019Berrick}, open-source expressive faces \cite{Faraj2020FaciallyExpressive}, and modular 3D-printed robotic heads \cite{Netzev2019ManyFacedRobot}. However, many of these platforms focus mainly on mechanical expressiveness or platform design, without deeply integrating visual attention estimation, semantic scene interpretation, and state-based social regulation.

Recent advances in vision--language models offer new possibilities for contextual interpretation in social robotics. Janssens et al. \cite{Janssens2025MultimodalSocialConversations} highlight the potential of VLMs for multimodal social interaction, although their integration into real-time embodied systems remains challenging. In this context, VLMs are particularly relevant because they can provide semantic labels that go beyond geometric cues, such as distinguishing whether a user is attending to the robot, looking at a phone, or directing attention elsewhere. However, their lower update rate and higher inference latency make them less suitable as direct control signals in time-sensitive HRI loops.

In this context, the present work contributes an applied case study that combines a fast geometric attention layer, an independent semantic VLM observer, and an expressive state machine in a physically accessible robotic head based on the InMoov ecosystem \cite{Langevin2012InMoov}. Rather than replacing geometric attention estimation with semantic interpretation, the proposed pipeline studies their complementary roles: the geometric layer provides high-frequency temporal control, while the VLM provides lower-frequency contextual interpretation for comparison and ambiguity analysis.

\section{Methodology}

\begin{figure}[h]
    \centering
    \includegraphics[width=\linewidth]{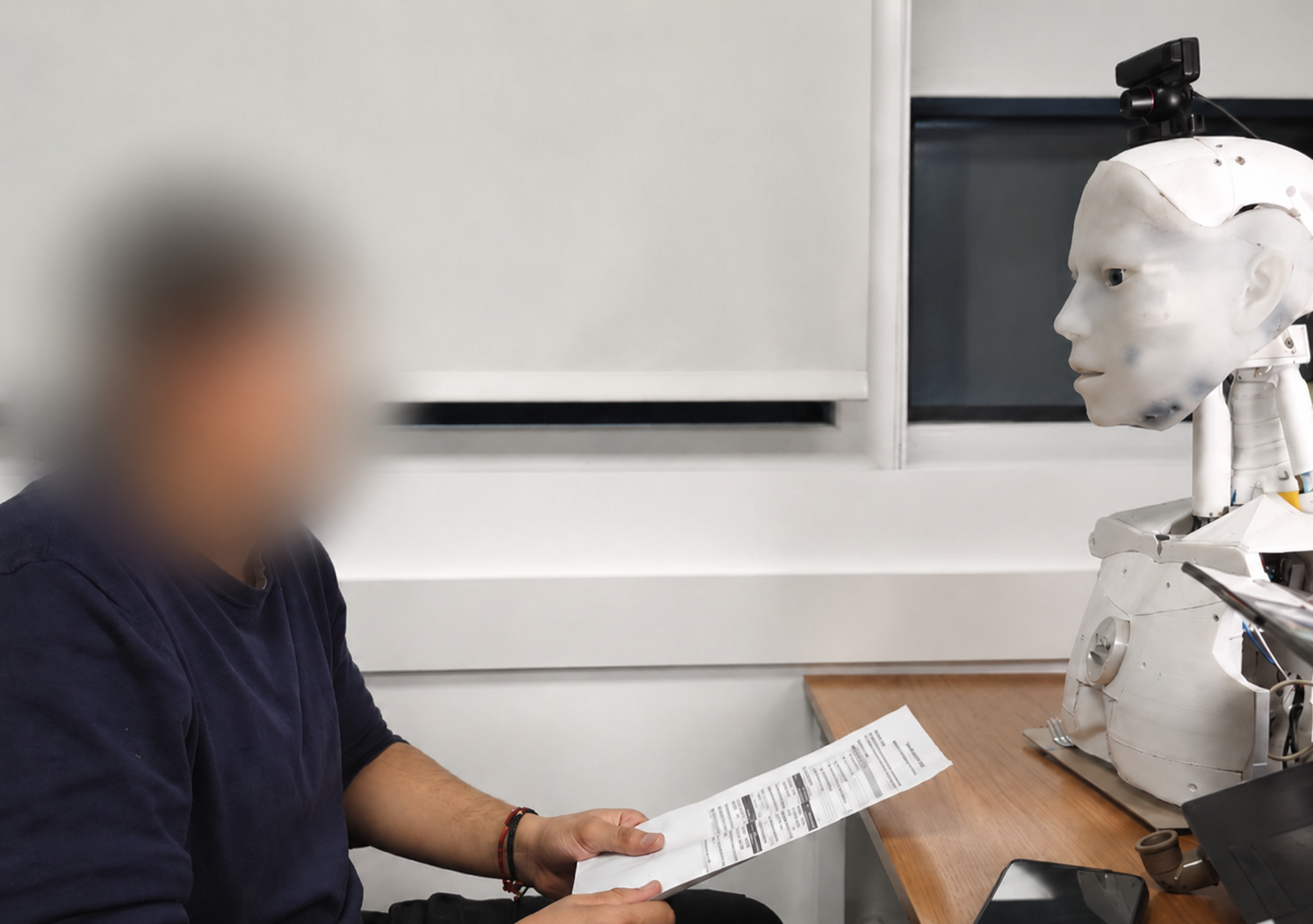}
    \caption{Face-to-face experimental setup with anonymized participant.}
    \label{fig:experimental_setup_real}
\end{figure}

\begin{figure*}[t]
    \centering
    \includegraphics[width=\textwidth]{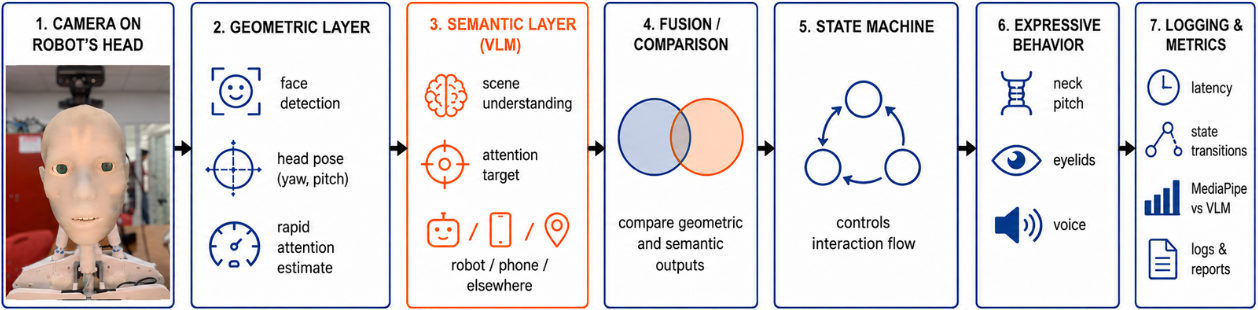}
    \caption{Hybrid attention-estimation pipeline with parallel geometric and semantic perception layers.}
    \label{fig:system_pipeline}
\end{figure*}

The proposed methodology evaluates a hybrid visual attention estimation pipeline for human--robot interaction, implemented on an expressive robotic head. The system transforms user attentional cues into discrete social states and observable adaptive behaviors. Fig.~\ref{fig:experimental_setup_real} shows the physical interaction setup, while Fig.~\ref{fig:system_pipeline} summarizes the complete pipeline.

\subsection{Robotic Platform and Perception Pipeline}

The experimental platform consists of an expressive robotic head based on the InMoov ecosystem, adapted for HRI experimentation. The current implementation uses 16 servo-driven facial channels for expressive motion, including jaw, eyelids, eyebrows, cheeks, and eye-orientation mechanisms, together with neck motion for head posture. A camera was mounted on the upper part of the head to provide a stable egocentric view of the interaction scene. This placement was chosen instead of embedding cameras inside the eyes in order to preserve the mechanical structure of the head, simplify calibration, and avoid interfering with the expressive eye mechanisms.

The perception and interaction pipeline runs on an NVIDIA Jetson Orin Nano integrated into the robotic platform. The camera stream is processed by two parallel perception layers: a geometric perception layer for fast attention estimation and state regulation, and a semantic layer for independent contextual interpretation. The geometric layer drives the state machine, while the semantic layer is logged and compared against the geometric estimate without directly controlling the robot.

The geometric perception layer was implemented using MediaPipe for face detection and head-pose estimation \cite{Lugaresi2019MediaPipe}. Hereafter, we refer to it as the geometric perception layer, since the relevant concept is the fast estimation of user orientation rather than the specific software tool. The layer estimates yaw and pitch angles from the participant's head pose and classifies the user as attentive when both angles remain within predefined thresholds for the interaction setup.

The thresholds were empirically defined during pilot calibration from the yaw and pitch ranges observed when participants faced the robot, and were selected to separate direct attention from clear head rotations toward the phone or elsewhere. This follows previous HRI works in which head orientation is used as a practical proxy for attentional direction \cite{Veronese2017ProbabilisticMapping,Li2012VisionBasedAttention,Das2012VisionBasedAttentionControl}. To avoid false transitions caused by frame-level noise, attention and inattention are confirmed only when the corresponding signal is sustained for a predefined temporal window.

The semantic layer is based on a vision--language model (VLM), implemented locally through Ollama/LLaVA-Phi3. In this pipeline, the VLM acts as an \textit{Independent Semantic Observer}: it does not receive the robot's internal state, geometric perception outputs, attention flags, pose angles, or annotated overlays. Instead, it receives only a clean raw frame captured from the robot's camera. Its output consists of a brief scene summary, a person-presence flag, an attention target, and a confidence level. The semantic attention targets considered in this study are \textit{robot}, \textit{phone}, and \textit{elsewhere}. The VLM does not directly trigger state transitions or control the robot; instead, its output is used for semantic comparison and post-hoc analysis of agreement and divergence with the geometric perception layer.

\subsection{State Machine and Experimental Protocol}

The system translates perceptual information into a discrete social state machine. The main states are \textit{sleeping}, \textit{person\_detected}, \textit{attention\_detected}, \textit{interacting}, \textit{user\_distracted}, \textit{waiting}, \textit{resume\_interaction}, and \textit{finished}. The robot starts in a resting state and transitions to active interaction when a participant is detected and visual attention is established. When sustained inattention is detected in the adaptive condition, the system enters a waiting state. If attention is recovered, the interaction resumes; otherwise, the robot returns to rest.

\begin{figure}[h]
    \centering
    \includegraphics[width=\linewidth]{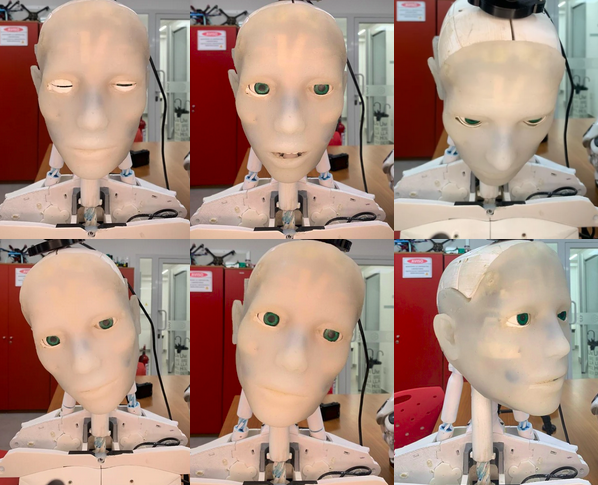}
    \caption{Representative expressive states of the robotic head.}
    \label{fig:state_machine}
\end{figure}

The study was structured around two conditions and two scenarios. In \textbf{Condition A} (baseline), the robot does not adaptively pause or resume the interaction in response to user distraction. In \textbf{Condition B} (adaptive), the robot monitors attention, pauses the interaction when sustained inattention is detected, waits for attention recovery, and resumes the interaction when attention is re-established. The two scenarios were: \textbf{Scenario 1}, continuous attention toward the robot, and \textbf{Scenario 2}, interrupted attention, for example by looking at a phone or looking away before recovering attention.

The evaluation involved 10 participants recruited by convenience sampling from the university environment. This sample was selected for an initial controlled, system-oriented evaluation with participants able to follow the experimental instructions. Formal demographic variables such as age and gender were not used as experimental factors in this study. Each participant completed four trials: A1, A2, B1, and B2, resulting in 40 valid trials.

\subsection{Data Logging and Analysis}

During each trial, the system records frame-by-frame logs and trial-level summaries. The logs include timestamps, participant and trial identifiers, experimental condition, scenario, current state, previous state, geometric perception outputs, semantic outputs, audio state, event markers, and relevant timing information. The trial-level summaries store the main event timestamps, including person detection, attention detection, speech start, distraction detection, pause trigger, attention recovery, resume trigger, timeout, and return to rest.

Since the internal trial identifier was reused across several files, each trial was uniquely identified during analysis using a normalized combination of participant identifier, condition, scenario, and timestamp. Whitespace in participant and trial identifiers was removed before aggregation. This normalization avoided incorrect grouping and resulted in 40 valid trials.

The analysis focused on reconstructing the interaction chain from person detection and attention detection to interaction start, state transition, and robot response. First, the start of each trial was validated by checking whether the robot detected the participant, detected attention, and started the spoken interaction. Second, protocol outcomes were computed for each condition and scenario, including pause and resume events. Third, geometric and semantic outputs were compared through a contingency matrix. Geometric outputs were grouped into \textit{looking}, \textit{not looking}, and \textit{no face}, and compared against the semantic targets \textit{robot}, \textit{phone}, and \textit{elsewhere}. Divergent cases were marked as ambiguous, not as direct errors, since no manual frame-level ground truth was annotated.

Finally, adaptive response timing was computed from the trial-level event timestamps. For adaptive trials, the pause latency was defined as

\begin{equation}
L_{pause}=t_{p}-t_{d},
\label{ec:pause_latency}
\end{equation}

where $t_{d}$ is the time of confirmed distraction and $t_{p}$ is the pause trigger time. This value does not represent the full latency from the participant's physical gaze shift to detection, since the exact onset of human distraction was not manually annotated.

\FloatBarrier

\section{Results and Discussion}

The experimental evaluation was conducted with 10 participants, each completing four trials: A1, A2, B1, and B2. This resulted in 40 valid trials. The analysis focused on three aspects: the reliability of the interaction start sequence, the behavior of the adaptive condition, and the relationship between the geometric perception layer and the semantic layer.

\subsection{Interaction Start and Protocol Outcomes}

Across all 40 trials, the robot detected the participant, detected visual attention, and started the spoken interaction in every case. The average time from trial start to person detection was $1.73 \pm 1.46$ s, attention detection occurred after $4.74 \pm 4.36$ s, and the spoken interaction started after $8.57 \pm 3.99$ s. These results indicate that the activation sequence was stable across participants and experimental conditions, although the time required to establish attention varied across trials.

Table~\ref{tab:interaction_start_validation} summarizes the interaction start validation results, while Table~\ref{tab:protocol_outcomes} reports the protocol outcomes by experimental condition. A1 and A2 are reported as baseline trials to verify that no adaptive pause or resume behavior was produced when the adaptive response was disabled. Therefore, the adaptive behavior should be interpreted mainly from B1 and B2, where the attention-driven pause/resume mechanism was active.


In B2, the robot triggered a pause in all 10 trials and resumed the interaction in 9 out of 10 trials. In B1, pause and resume events occurred in 3 out of 10 trials, although no planned distraction was introduced. This suggests that the attention-driven state machine can react to spontaneous or unintended attention losses, while also indicating that temporal thresholds should be further refined in future work.

\begin{table}[H]
\caption{Interaction start validation results.}
\label{tab:interaction_start_validation}
\centering
\begin{tabular}{lc}
\hline
\textbf{Metric} & \textbf{Value} \\
\hline
Participants & 10 \\
Valid trials & 40 \\
Person detected & 40/40 \\
Attention detected & 40/40 \\
Interaction started & 40/40 \\
Time to person detection & $1.73 \pm 1.46$ s \\
Time to attention detection & $4.74 \pm 4.36$ s \\
Time to speech start & $8.57 \pm 3.99$ s \\
\hline
\end{tabular}
\end{table}

\begin{table}[H]
\caption{Protocol outcomes by experimental condition.}
\label{tab:protocol_outcomes}
\centering
\begin{tabular}{lcccc}
\hline
\textbf{Case} & \textbf{Mode} & \textbf{Start} & \textbf{Pause} & \textbf{Resume} \\
\hline
A1 & Baseline & 10/10 & 0/10 & 0/10 \\
A2 & Baseline & 10/10 & 0/10 & 0/10 \\
B1 & Adaptive & 10/10 & 3/10 & 3/10 \\
B2 & Adaptive & 10/10 & 10/10 & 9/10 \\
\hline
\multicolumn{5}{l}{\footnotesize Start: successful interaction initiation.} \\
\multicolumn{5}{l}{\footnotesize Pause: successful adaptive interruption.} \\
\multicolumn{5}{l}{\footnotesize Resume: successful recovery after interruption.} \\
\end{tabular}
\end{table}


\subsection{Adaptive Response Timing}

For adaptive trials in which a pause occurred, the response latency was computed as the interval between the internally confirmed distracted state and the pause trigger. This metric does not represent the complete delay from the participant's physical gaze shift to system detection, since the exact onset of human distraction was not manually annotated. Instead, it measures how quickly the robot reacted once the state machine had confirmed sustained inattention.

In B1, the mean confirmed-distraction-to-pause latency was $33.0 \pm 1.0$ ms. In B2, the corresponding mean latency was $33.2 \pm 1.5$ ms. The mean pause-to-resume interval was $2800.3 \pm 985.0$ ms in B1 and $3976.6 \pm 1014.4$ ms in B2. These results indicate that the pause action was triggered immediately after internal confirmation, while the pause-to-resume interval reflected the time required for the participant to recover attention and for the system to confirm the recovery.

\begin{table}[H]
\caption{Adaptive response timing.}
\label{tab:adaptive_response_timing}
\centering
\begin{tabular}{lcc}
\hline
\textbf{Case} & \textbf{Pause latency} & \textbf{Pause--resume interval} \\
\hline
B1 & $33.0 \pm 1.0$ ms & $2800.3 \pm 985.0$ ms \\
B2 & $33.2 \pm 1.5$ ms & $3976.6 \pm 1014.4$ ms \\
\hline
\end{tabular}
\end{table}


\subsection{Geometric-Semantic Comparison}

A total of 464 valid semantic updates were compared with the corresponding geometric perception estimates. The semantic layer classified 285 updates as \textit{robot}, 92 as \textit{phone}, and 87 as \textit{elsewhere}. This corresponds to 61.4\%, 19.8\%, and 18.8\% of the valid semantic updates, respectively. Therefore, the semantic layer did not collapse into a single category and produced labels associated with direct attention to the robot, phone-related distraction, and attention elsewhere.

\begin{figure}[t]
    \centering
    \includegraphics[width=\linewidth]{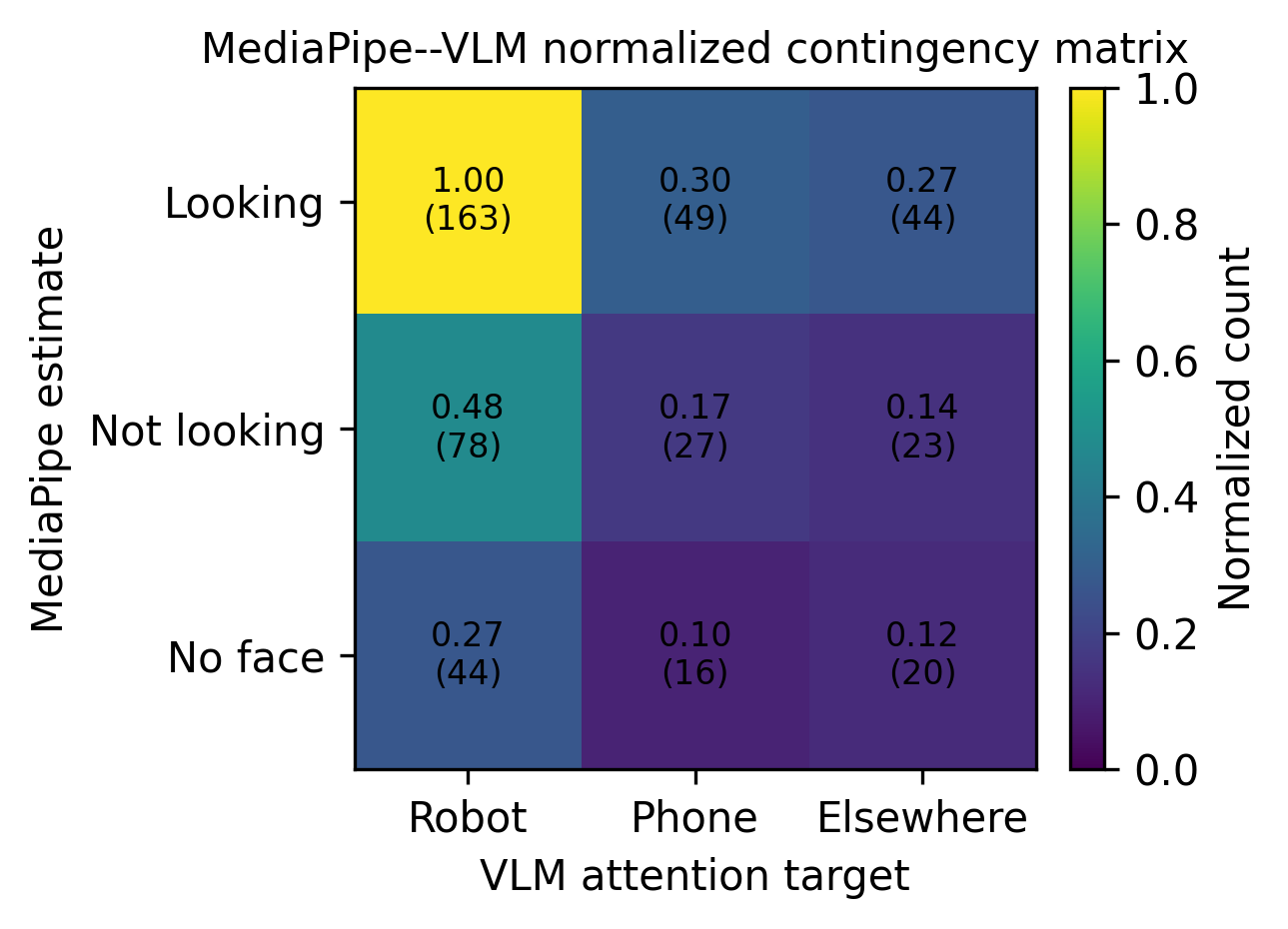}
    \caption{Normalized geometric--semantic contingency matrix. Values are normalized by the maximum cell count; absolute counts are shown in parentheses.}
    \label{fig:mediapipe_vlm_matrix}
\end{figure}

Fig.~\ref{fig:mediapipe_vlm_matrix} shows the contingency matrix between the geometric and semantic outputs. The matrix reveals several cases where the two perception layers provided non-equivalent descriptions of the same interaction moment. Using the predefined divergence rules, 251 out of 464 valid comparisons were identified as ambiguous or divergent cases. These cases were not interpreted as direct classification errors, since no manual frame-level ground truth was annotated. Instead, they represent moments where geometric attention estimation and semantic scene interpretation did not provide equivalent descriptions of the interaction.

This result suggests that head-pose-based attention estimation and semantic scene interpretation should not be treated as equivalent signals, especially in embodied HRI scenarios where the user's visual orientation and contextual target may not always coincide. This supports the use of a hybrid perception pipeline in which a fast geometric layer is complemented by an independent semantic observer.

\FloatBarrier

\section{Conclusion}

This paper presented an applied case study on hybrid visual attention estimation for HRI using an expressive robotic head based on the InMoov ecosystem. The proposed pipeline combines a fast geometric perception layer with an independent VLM-based semantic observer, connected through a state machine that regulates adaptive interaction behavior.

The system was evaluated with 10 participants across 40 trials. Results showed that the robot successfully detected the participant, detected attention, and started the interaction in all trials. In the adaptive distraction scenario, the robot paused the interaction in 10 out of 10 trials and resumed in 9 out of 10 trials. The geometric--semantic comparison showed that semantic labels were not fully redundant with geometric attention estimates, supporting the use of parallel perception layers in attention-aware HRI.

The main limitations of this study are the controlled experimental setup, the reduced number of participants, and the absence of manual frame-level ground truth for attention labels. Future work will focus on refining attention thresholds, adding manual annotation for stronger validation, and evaluating how adaptive expressive behavior affects user perception and interaction quality.

\section*{Acknowledgment}

The authors would like to thank the Robotics and Artificial Intelligence Laboratory for supporting this work, and Gael Langevin, creator of the InMoov project, for making available the open-source humanoid platform on which the expressive robotic head used in this study was based.


\bibliographystyle{IEEEtran}
\bibliography{bibliography/IEEEexample}

\end{document}